\documentclass[sigplan]{acmart}

\usepackage{booktabs} 
\usepackage{times}
\usepackage{latexsym}
\usepackage{graphicx}
\usepackage{url}
\usepackage{algorithm} 
\usepackage{algpseudocode}
\usepackage{subfig}
\usepackage{comment}
\usepackage[export]{adjustbox}
\usepackage{upgreek}
\usepackage{multirow}
\usepackage{color}
\usepackage{array}
\usepackage{comment}
\usepackage{multirow}

\newcommand{\etal}{et al.\ } 

\acmConference[]{}{}{}

\begin{document}
\title[Joint Multi-Domain Learning for ASAG]{Joint Multi-Domain Learning for Automatic Short Answer Grading}

\author{Swarnadeep Saha}
\affiliation{%
  \institution{IBM Research - India}
}
\email{swarnads@in.ibm.com}

\author{Tejas I. Dhamecha}
\affiliation{%
  \institution{IBM Research - India}
}
\email{tidhamecha@in.ibm.com}

\author{Smit Marvaniya}
\affiliation{%
  \institution{IBM Research - India}
}
\email{smarvani@in.ibm.com}

\author{Peter Foltz}
\affiliation{%
  \institution{Pearson}
}
\email{peter.foltz@pearson.com}

\author{Renuka Sindhgatta}
\affiliation{%
  \institution{Queensland University of Technology}
}
\email{renuka.sr@qut.edu.au}

\author{Bikram Sengupta}
\affiliation{%
  \institution{Anudip Foundation}
}
\email{bikramsengupta@gmail.com}

\renewcommand{\shortauthors}{Saha et al.}

\begin{abstract}
One of the fundamental challenges towards building any intelligent tutoring system is its ability to automatically grade short student answers. A typical automatic short answer grading system (ASAG) grades student answers across multiple domains (or subjects). Grading student answers requires building a supervised machine learning model that evaluates the similarity of the student answer with the reference answer(s).  
We observe that unlike typical textual similarity or entailment tasks, the notion of similarity is not universal here.
On one hand, para-phrasal constructs of the language can indicate similarity independent of the domain. On the other hand, two words, or phrases, that are not strict synonyms of each other, might mean the same in certain domains. Building on this observation, we propose JMD-ASAG, the first joint multi-domain deep learning architecture for automatic short answer grading that performs domain adaptation by learning generic and domain-specific aspects from the limited domain-wise training data. JMD-ASAG not only learns the domain-specific characteristics but also overcomes the dependence on a large corpus by learning the generic characteristics from the task-specific data itself. On a large-scale industry dataset and a benchmarking dataset, we show that our model performs significantly better than existing techniques which either learn domain-specific models or adapt a generic similarity scoring model from a large corpus. Further, on the benchmarking dataset, we report state-of-the-art results against all existing non-neural and neural models.

\end{abstract}

%
%


\maketitle

\section{Introduction}
\begin{figure}
    \centering
    \includegraphics[width=\linewidth]{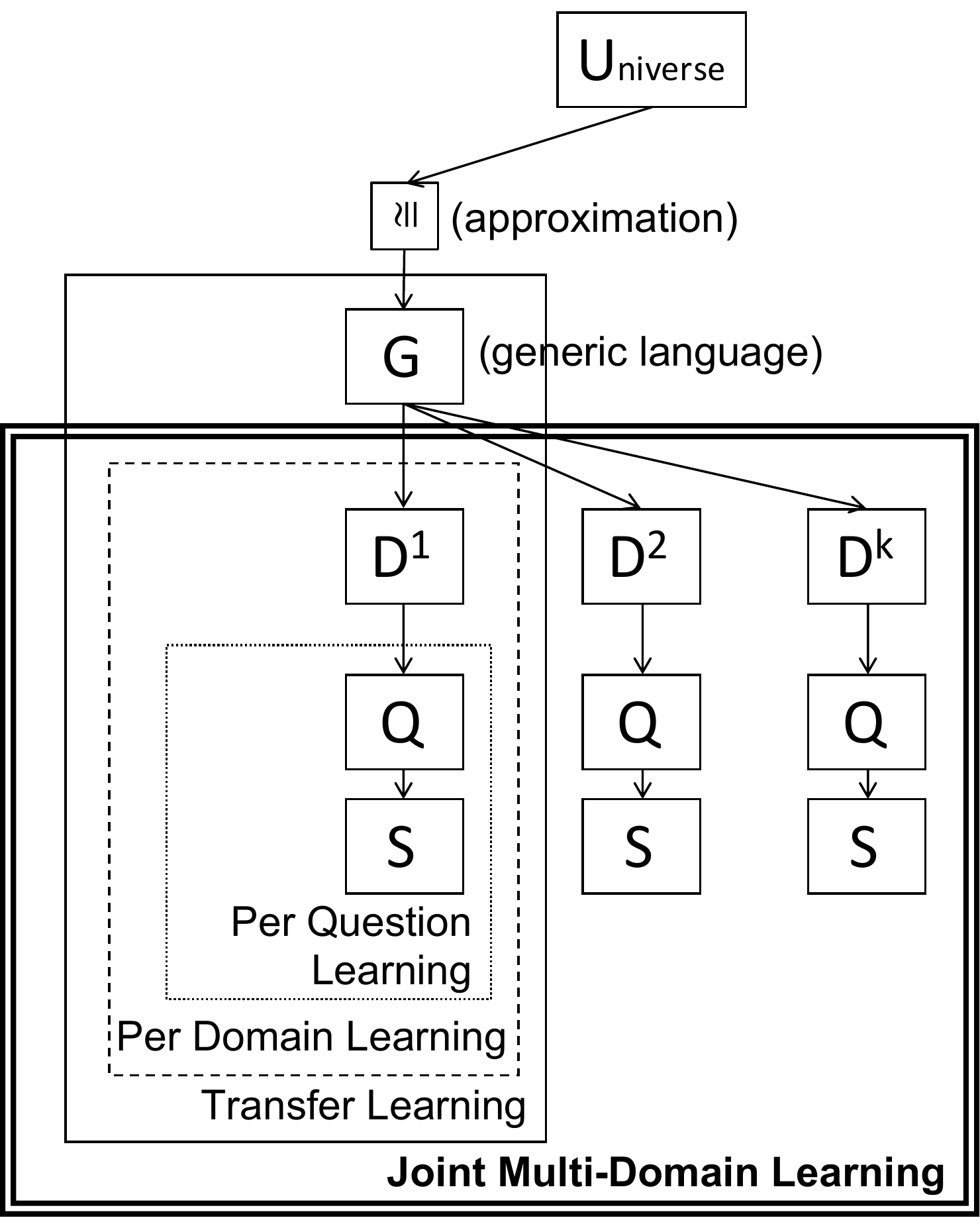}
    \caption{Various existing approaches for automatic short-answer grading can be broadly categorized in 1) per question learning, 2) per domain learning, and 3) transfer learning. The proposed approach involving joint multi-domain learning removes need for a large generic language corpus, and, compensates for it by joint learning of domain-specific and generic classifiers. Accompanying formulation details are in Table \ref{tab:awesome_2}.}
    \label{fig:awesome}
\end{figure}
\begin{table*}
    \centering
    \begin{tabular}{|c|c|p{6cm}|}
    \hline
         Model & Classifier & \multicolumn{1}{c|}{Description}\\\hline\hline
         $Q \rightarrow R$ & $f^Q(R, S)$ & per question modelling\\\hline
         $D \rightarrow (Q,R)$ & $f^D(Q,R,S)$ & per domain modelling\\\hline
         $D^G \cong D^S \implies$ $(D \rightarrow (Q,R))$ & $f^G\implies f^D(Q,R,S)$ & transfer or adapt from generic source domain to task-specific target domain\\\hline
         $D^G \cong (D^1 \oplus D^2 \oplus \cdots \oplus D^k) \implies (D^i \rightarrow (Q,R))$ & $ f^{D^i}(Q,R,S)$ and $f^G(Q,R,S)$ & joint multi-domain learning\\\hline
    \end{tabular}
    \caption{An illustration of various approaches to model short answer grading problem. $D$, $Q$, $R$, and $S$, represent domain, question, reference answer, and student answer, respectively. $D^G$, $D^S$, and $D^i$, represent generic, source, and $i^{th}$ task domains, respectively. In transfer learning school of thought, model learned for generic task (e.g. natural language inference) is adapted to a specific task. In the proposed joint multi-domain learning approach, a model capturing generic language characteristic ($f^G$) is jointly learned with domain-specific models ($f^{D^i}$) without requiring large generic source ($D^S$) corpus.} \label{tab:awesome_2}
\end{table*}

Automatically grading short student answers is critical for building Socratic intelligent tutoring systems \cite{rose2001comparative}.
In general, computer-aided assessment systems are particularly useful because grading by humans can become monotonous and tedious \cite{kumar2017earth}. 
Formally, the problem of Automatic Short Answer Grading (ASAG) is defined as one where for a given question, a short student answer (typically, 1-2 sentences long) is graded against the reference answer(s).

Figure \ref{fig:awesome} and Table \ref{tab:awesome_2} illustrate various strategies for ASAG. One of the strategies is to assume that for every question $Q$, a variety of reference answers $R$ is available during training, i.e. the testing scenario is unseen-answer only. Under this assumptions, a classifier $f^Q$ can be trained per question. However, such approaches cannot generalize to unseen-questions and unseen-domains.

To make an approach generalizable to unseen-questions, one can learn a classifier $f^D$ per domain. Each subject (e.g. Primary Science) can be treated as a domain. Such approaches alleviate the need for large number of training answer variants for each question. Grading of a student answer is performed conditional to the question and the reference answer(s). Traditionally, these supervised approaches for ASAG use hand-crafted features to model the similarity between the reference answers and the student answers \cite{Mohler,jimenez2013softcardinality,SultanSS16}.  Such techniques succeed in capturing domain specific characteristics; however, their performance is sensitive to feature engineering. Deep learning (DL) approaches can mitigate the need for hand-crafting features, but rely heavily on availability of large data.

Automatic short-answer grading task lacks large scale data to efficiently train existing architectures of DL models. 
In absence of domain and task specific large scale data, 
transfer learning is explored \cite{bowman2015large,moutransfer,conneau2017supervised,gensem}.
It builds on the intuition that a source domain $D^S$ and corresponding generic task can help learn embeddings (or classifier) $f^G$ that approximates universal characteristics of language. Such a model is then transferred to the task-specific domain to obtain the final classifier $f^D$; either by fine-tuning generic embeddings or by learning a
task-specific classifier over generic embeddings. However, we believe that there is a scope for significant improvement in this strategy under certain scenarios.

We propose a joint multi-domain learning approach for short-answer grading. The proposed modelling does not assume availability of any other data beyond the corpus consisting of multiple task-specific domains ($D^1$-$D^k$). It jointly learns domain specific classifiers $f^{D^i}$ and a generic classifier $f^G$. Particularly, we believe that this strategy can be very helpful under certain scenarios:


\begin{table}[!t]
\small
\begin{tabular}{|l|p{6cm}|}
\hline
Question & Darla tied one end of a string around a doorknob and held the other end in her hand. When she plucked the string (pulled and let go quickly) she heard a sound. How would the pitch change if Darla pulled the string tighter?\\\hline
Ref. answer & When the string is tighter, the pitch will be higher. \\\hline
Std. answer & The pitch would be higher if she pulled it really tight \\\hline
 \multicolumn{2}{|c|}{\texttt{(When X, Y)} = \texttt{(Y, if X)}}\\\hline
\multicolumn{2}{c}{(a)  Generic textual characteristics}\\\hline

Question & Lee has an object he wants to test to see if it is an insulator or a conductor. He is going to use the circuit you see in the picture. Explain how he can use the circuit to test the object. \\\hline
Ref. answer & If the motor runs, the object is a conductor. \\\hline
Std. answer & He could know if it works.\\\hline
\multicolumn{2}{|c|}{\texttt{(X runs)} = \texttt{(X works)}}\\\hline
\multicolumn{2}{c}{(b)  Domain-specific characteristics}\\
\end{tabular}
\caption{ Two examples from the SemEval-2013 dataset illustrating the importance of generic and domain-specific characteristics in ASAG.}\label{tab:example}
\end{table}

\begin{enumerate}
\item If the end task (e.g. ASAG) is pre-defined, it may be well-suited to train the task-specific model, as compared to transferring or fine-tuning. Effectively, the problem boils down to learning from the limited task-specific training data.

\item If within the pre-defined task, there exists specific domains (e.g. short answer grading for Psychology and Criminology), an adaption of learning across them may be more effective.

\end{enumerate}

Note that these suggestions also help reduce the dependence on a large corpus. The former learns only the task-specific aspects rather than the language itself, and the latter adapts to the domains by learning both the generic and domain-specific characteristics within the task.


We find that these scenarios are often prevalent in the task of ASAG for intelligent tutoring systems; where, it is likely to have various domain-specific smaller corpora for individual subjects. Although it is hard to train DL models individually for each domain due to their limited sizes, put together, the corpora from various domains can provide sufficient view of the language understanding. Consider the examples in Table \ref{tab:example}. In the first example, to successfully match the student answer with the reference answer, the model needs to understand a grammatical construct that \texttt{When X, Y} is paraphrase to \texttt{Y, if X}. In order to learn this construct, the training set should contain examples with this syntax; but may not necessarily be from the same domain. In the second example, the system is required to learn that \texttt{X runs} and \texttt{X works} mean the same in the particular domain (in this case, electrical). To successfully understand constructs like these, it is required to have domain-specific training data. Building upon these intuitions, we make the following contributions.


\begin{itemize}
\item We motivate the need for a joint multi-domain model for ASAG as unlike typical short text similarity tasks, the meaning of similarity in ASAG can vary across domains. Our examples show the domain-specific as well as generic aspects of similarity.
\item We propose a novel Joint Multi-Domain neural model for ASAG (JMD-ASAG) that learns generic and domain-specific aspects simultaneously. It achieves this by utilizing multiple domain-specific corpora, and without requiring a large generic corpus.
\item To evaluate the hypothesis of utilizing task-specific corpus, we show the effectiveness of JMD-ASAG compared to a state-of-the-art method that performs transfer learning from a large corpus. 
\item We compare JMD-ASAG with its generic and domain-specific components on a benchmarking dataset and a proprietary industry dataset. 
It outperforms both and also achieves improved results on the benchmarking dataset compared to various state-of-the-art non-neural and neural models.
\end{itemize}

\begin{figure*}
\subfloat[][Encoder. \label{fig:encoder}]{\includegraphics[width=0.1\textwidth,valign=c]{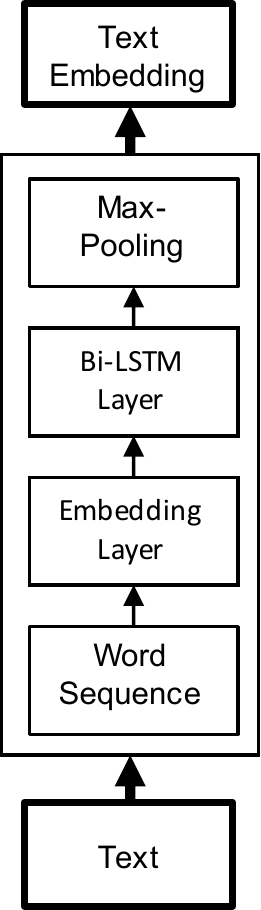}
\vphantom{\includegraphics[width=0.5\textwidth, valign=c]{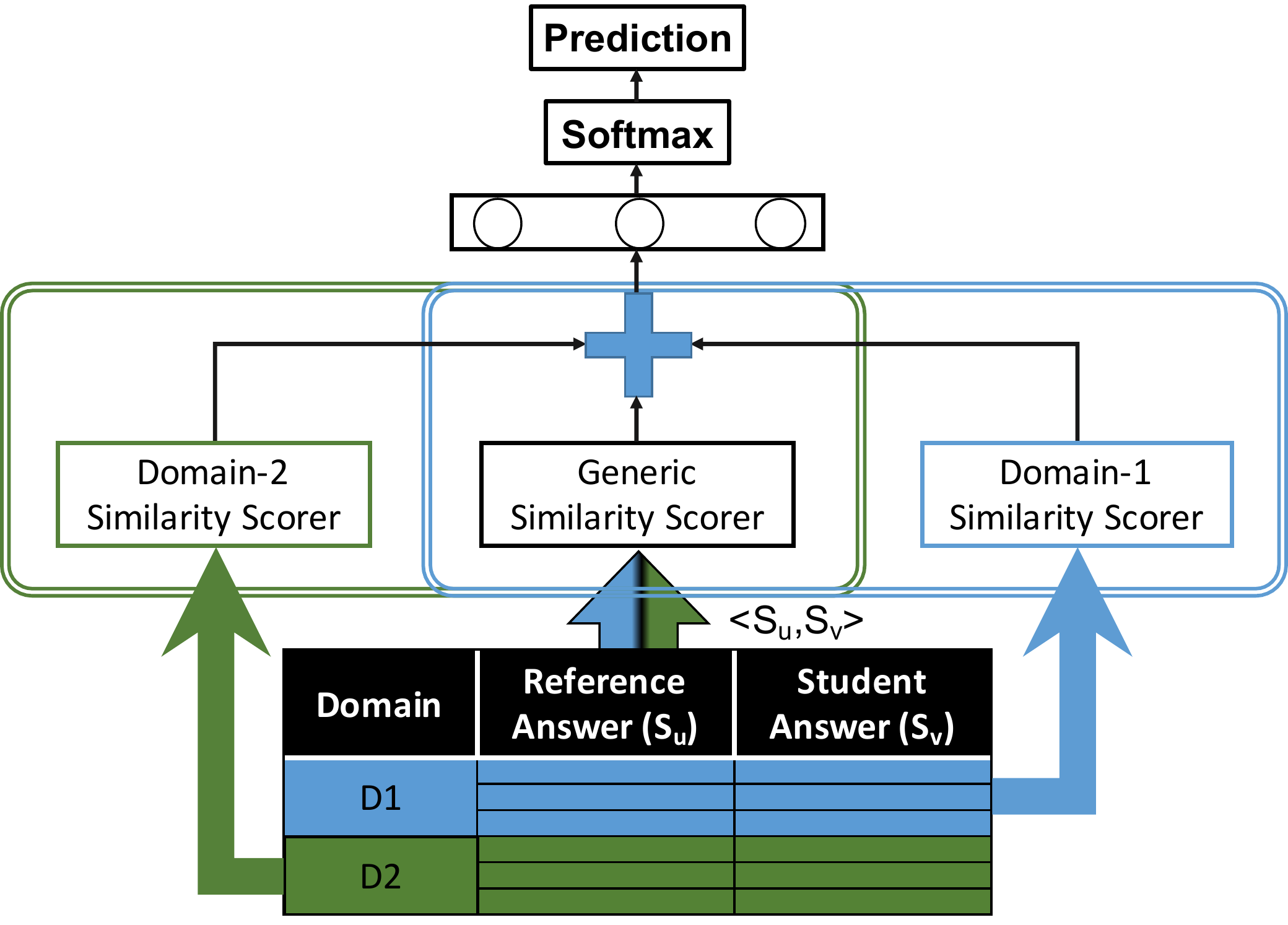}}}\quad\vrule\quad
\subfloat[][Similarity Scorer. \label{fig:scorer}]{\includegraphics[width=0.25\textwidth,valign=c]{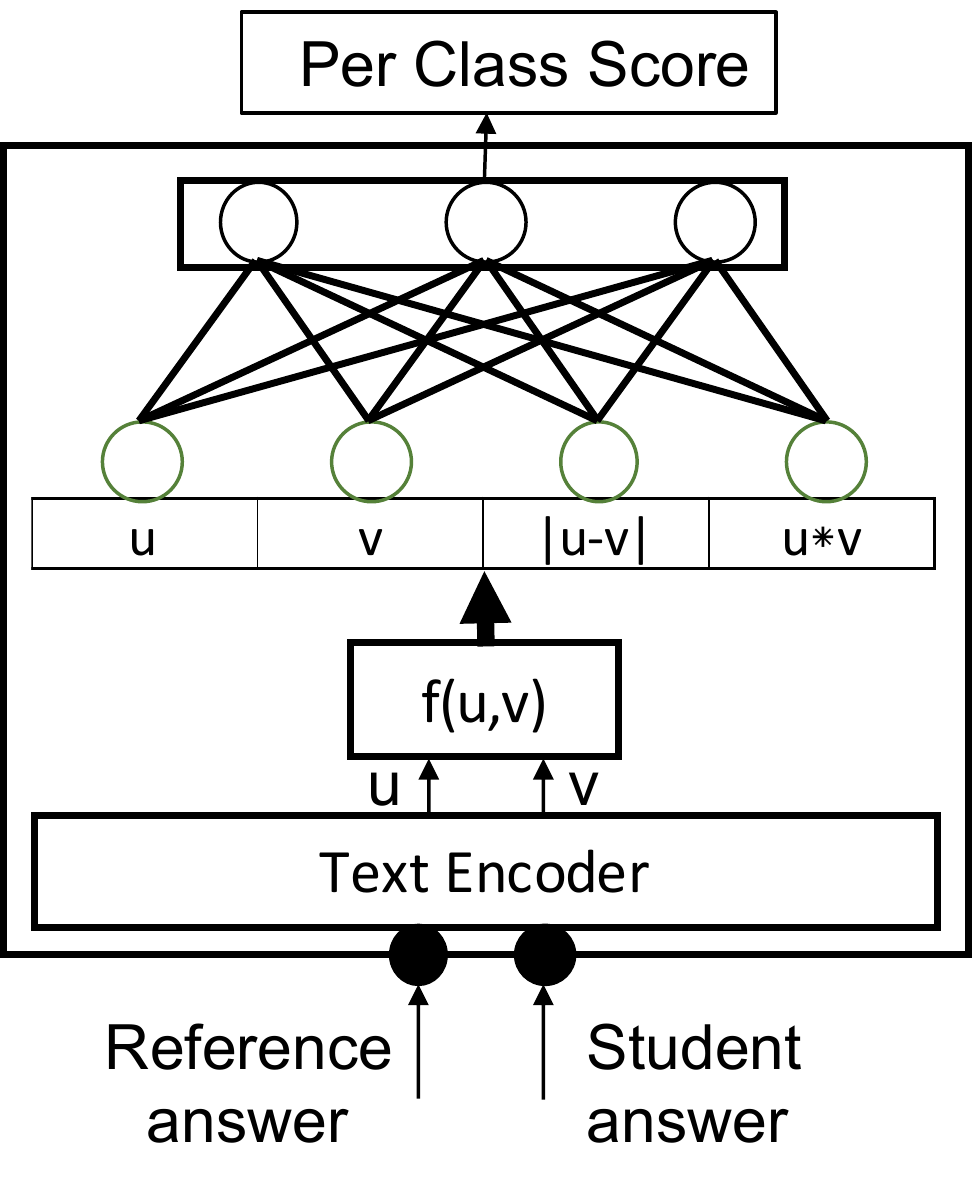}
\vphantom{\includegraphics[width=0.5\textwidth, valign=c]{block2.pdf}}}\quad\vrule\quad
\subfloat[][Overall architecture. \label{fig:block}]{\includegraphics[width=0.5\textwidth, valign=c]{block2.pdf}}

\caption{Individual components and overall architecture of Joint Multi-Domain ASAG.\label{fig:approach}}
\end{figure*}

\section{Related Work}
This research is positioned at the intersection of domain adaptation and its utility to improve ASAG. Following is a broad overview of related works in these fields of research.


\subsection{Automatic Short Answer Grading}
Traditional approaches of ASAG range from applying manually generated or automated patterns \cite{mitchell2002towards,nielsen2009recognizing,sukkarieh2004auto,ramachandran2015identifying} to using hand-crafted features, that include graph alignment features \cite{Mohler,SultanSS16}, n-gram features \cite{heilman2013ets}, softcardinality text overlap features \cite{jimenez2013softcardinality}, averaged word vector text similarity features \cite{SultanSS16} and other shallow lexical features \cite{ott2013comet,Mohler}. 

More recently, deep learning techniques have been explored -  \citet{riordan2017investigating} adapts the convolutional recurrent neural network, originally proposed by \citet{taghipour2016neural} for automated essay scoring and \citet{kumar2017earth} uses Earth Mover's Distance Pooling over Siamese BiLSTMs.
Among other approaches which view this problem as an application of semantic textual similarity, the most recent one, InferSent \cite{conneau2017supervised} uses a max pooled bidirectional LSTM network to learn universal sentence embeddings from the MultiNLI corpus \cite{williams2017broad}. These embeddings have been employed as features in conjunction with hand-crafted features by \citet{Saha2018} for ASAG.

\subsection{Neural Domain Adaptation}
Domain Adaptation, with or without neural networks, has been an active area of research for the past decade.
\citet{daume2007frustratingly} proposes a highly efficient domain adaptation method based on feature augmentation but one which considers mostly sparse binary-valued features. 
This is further extended by \citet{kim2016frustratingly} for dense real-valued features to facilitate usage in  neural networks. They use $k+1$ LSTMs where $k$ of them capture domain-specific information and one is useful for generic or global information. Other works on domain adaptation augment the $k$ domain-specific models with a domain-specific parameter \cite{alumae2013multi,tilk2014multi,ammar2016many} but unlike our work, do not have a generic component. Finally, \citet{Chen2018} propose a multinomial adversarial learning framework for multi-domain text classification but restricting themselves to tasks of sentiment classification only. Importantly, none of these neural models perform multi-domain learning for short text similarity which is particularly useful for ASAG as motivated before.

Neural domain adaptation is closely related to neural multi-task learning where one single architecture is developed to work across multiple related tasks. It has found applications in sequence tagging \cite{yang2016multi,sogaard2016deep}, semantic parsing \cite{peng2017deep} and pairwise sequence classification tasks \cite{augenstein2018multi}.
\citet{liu2017} employ adversarial learning for better separation of shared and private features on related text classification tasks. Finally, \citet{peng2016multi} combine domain adaptation and multi-task learning for sequence tagging. They propose an architecture where the BiLSTM embedding output is masked into $k+1$ parts, representing generic and domain-specific features but do not learn separate components for each of them.

\subsection{Domain Adaptation for ASAG}
Domain adaptation for ASAG has been a relatively less explored area thus far. 
Notably, \citet{heilman2013ets} propose domain adaptation for ASAG by applying \citet{daume2007frustratingly}'s feature augmentation method to create multiple copies of hand-crafted features. We directly compare against them in the experiments section.

To the best of our knowledge, neural domain adaptation for ASAG is unexplored in the literature. In this research, we propose a neural domain adaptation approach that explores multi-domain information in the context of ASAG.

\section{Method}
We propose JMD-ASAG, a Joint Multi-Domain neural network architecture for domain adaptation of ASAG.
We discuss our method in two parts - (1) the neural architecture of JMD-ASAG and (2) the training algorithm of JMD-ASAG.

\subsection{Neural Architecture}
The block diagram for the architecture is shown in Figure \ref{fig:approach}. For simplicity, it considers two domains but can be generalized to an arbitrary number of domains. We first consider the two key components of the model - (1) a Text Encoder (Figure \ref{fig:encoder}) and (2) a Text Similarity Scorer (Figure \ref{fig:scorer}). Later, we use them to build our overall model (Figure \ref{fig:block}).

\subsubsection{Text Encoder}
The text encoder provides dense feature representation of an input text (in this case, answer). We use bidirectional long short-term memory (BiLSTM) network \cite{hochreiter1997long} with max-pooling to encode the input answer, as detailed below.
We first embed each word in the answer using an embedding layer. The words are initialized with pre-trained word embeddings and are made trainable to reflect the domain and task dependent nature of the words. 
The sequence of words are then passed through a BiLSTM  layer to generate a sequence of hidden representations. Formally, for a sequence of $T$ words $\{w_t\}_{t=1,\ldots,T}$, the BiLSTM layer generates a sequence of $\{h_t\}$ vectors, where $h_t$ is the concatenation of a forward and a backward LSTM output:
\begin{eqnarray}
\{h_t^f\} &=& LSTM(w_1, w_2,\ldots,w_T) \nonumber\\
\{h_t^b\} &=& LSTM(w_T, w_{T-1},\ldots,w_1)\nonumber\\
h_t &=& [h_t^f, h_t^b]\ \ \ \  \forall t = 1,\ldots, T\nonumber
\end{eqnarray}
The hidden vectors $\{h_t\}$ are then converted into a single vector using max-pooling, which chooses the maximum value over each dimension of the hidden units. This fixed size vector is used as the vector representation for the input text. Overall, the text encoder can be treated as an operator $\mathcal{E}:\mbox{Text}\rightarrow\mathcal{R}^d$ that provides $d-$dimensional encoding for a given text.
Similar architectures for text encoders have been explored before, most notably by \cite{conneau2017supervised} for learning universal sentence embeddings.

\subsubsection{Text Similarity Scorer}
The text similarity scorer processes a reference answer ($R$) and a student answer ($S$) pair $\{R,S\}$ to generate class-wise scores. Their textual encodings are obtained using the aforementioned encoder as $\mathcal{E}(R)$  and $\mathcal{E}(S)$, respectively.
These encodings are used to compute the similarity feature representation $f$. It is formed by concatenating the (1) the reference answer encoding, (2) the student answer encoding, (3) their element-wise multiplication, and (4) their absolute difference.
\begin{eqnarray}
f = [\mathcal{E}(R),\  \mathcal{E}(S),\  \mathcal{E}(R)*\mathcal{E}(S),\  |\mathcal{E}(R)-\mathcal{E}(S)|]\nonumber
\end{eqnarray}
Note that the dimensionality of the feature $f $ is $4d$, where $d$ is the dimensionality of the encoding. 
The element-wise multiplication and the absolute difference components help capture the information gap between the reference answer and the student answer. 
Finally, the feature representation is transformed to class-wise similarity scores, by learning a dense layer ($W$).
\begin{eqnarray}
s = W^\prime f , \mbox{\ \ \ where }  W \in \mathcal{R}^{4d\times c} \nonumber
\end{eqnarray}
The $c-$dimensional output of the dense layer represents score for the answer pair's $\{R,S\}$ association to each of the $c$ classes. Overall, the text similarity scorer can be treated as an operator $\mathcal{S}:{\{\mbox{Std.Answer, Ref.Answer}\}\rightarrow}\mathcal{R}^c$, that computes class-wise scores for a given pair of student and reference answer.

\subsubsection{Overall Architecture}

For $k$ domains $\{D_d\}_{d=1,2,..k}$, JMD-ASAG's neural network architecture consists of $k+1$ text similarity scorers - $k$ domain-specific scorers $\{\mathcal{S}_d\}_{d=1,2,...,k}$ and one generic scorer $\mathcal{S}_g$.
For a sample $x$ belonging to the $d^{th}$ domain, its class-wise similarity score is obtained using its corresponding domain-specific scorer $\mathcal{S}_d$ and the generic scorer $\mathcal{S}_g$.
Their scores are added and finally converted to class-wise probabilities using a softmax function $\sigma$.
\begin{eqnarray}
 &\mathcal{P}(x)= \sigma\left(\mathcal{S}_d(x) + \mathcal{S}_g(x)\right), \mbox{ where }  x \in D_d\label{eq:combined}& \nonumber
\end{eqnarray}

Note that, each scorer has its own set of parameters. In other words, the parameters are not shared across the scorers. The generic scorer is called so because it is trained using data from all the domains and thus learns aspects generic or common to all of them (e.g. example 1 in Table \ref{tab:example}). The domain-specific ones are trained only using their corresponding domain's data and thus learn the domain-specific characteristics (e.g. example 2 in Table \ref{tab:example}). These components of the overall network enable it to learn the generic and domain-specific characteristics of a short answer grader from the task-specific data itself.
\begin{algorithm}[tp]
\caption{Training JMD-ASAG}\label{alg:euclid}
\begin{algorithmic}[1]

\Procedure{train\_model}{$domains$}
\State $k$ = len($domains$)
\State initialize $model$ 
\For{$e = 1$ to $num\_epochs$} 
    \For{$b = 1$ to $num\_batches$}
    	\For{$d = 1$ to $k$}
    		\State $batch$ = $b^{th}$ mini-batch of $domains[d]$
        	\State train\_on\_batch($model$, $batch$, $d$)
        \EndFor
    \EndFor
\EndFor
\EndProcedure

\Procedure{train\_on\_batch}{$model$, $batch$, $d$}
\State $S_g$ = $model$.GenericScorer($batch$)
\State $S_d$ = $model$.DomainScorer$[d]$($batch$)
\State Compute loss using $\sigma(S_g+S_d)$ and $batch.labels$
\State Back-propagate and update $model$
\EndProcedure
\end{algorithmic}
\end{algorithm}

\subsection{Training Algorithm}
We train JMD-ASAG using algorithm \ref{alg:euclid}. In every epoch, we generate batches by iterating over all the domains in one particular order. Note that the domain changes after every batch. 
In the architecture, the generic scorer $S_g$ is trained in each batch; whereas, depending on the domain $D_d$ of the batch, only the corresponding domain-specific scorer $S_d$ is trained. 
As part of the experiments, we explore other methods of training JMD-ASAG as well and evaluate their performances compared to the proposed one.

\section{Experiments}
In this section, we first demonstrate the effectiveness of the proposed JMD-ASAG on two datasets - (1) a proprietary large-scale industry dataset and (2) SemEval-2013 dataset \cite{Dzikovska}. 
For both the datasets, we compare our model with:
\begin{itemize}
\item \textbf{Transfer Learning:} We follow the \emph{learn universal and transfer} methodology suggested by \citet{conneau2017supervised} for transferring universal sentence embeddings. We generate embeddings for the reference answer and the student answer using their pre-trained BiLSTM with max-pooling network model\footnote{\texttt{infersent.allnli.pickle} model shipped with InferSent code is used.}, trained on the 430K sentence pairs of MultiNLI \cite{williams2017broad}.
These embeddings are used to compute the feature representation formed by concatenating their element-wise multiplication and absolute difference. Finally, we transfer these features for the task of ASAG using two configurations.
\begin{itemize}
\item \textbf{Generic Transfer Learning (GTrL): } We train one multinomial logistic regression model on the entire training set, formed by the combination of the training data from all the domains. The model is subsequently tested on each of the domains individually.
\item \textbf{Domain-specific Transfer Learning (DTrL): } We train multiple multinomial logistic regression models, one for each domain and subsequently test each of them on the corresponding domain only.
\end{itemize}
	


\item \textbf{Task-specific Learning: } As part of task-specific learning, we perform ablated comparisons with the generic and the domain-specific components of JMD-ASAG. Specifically, we compare with the following two configurations.

\begin{table*}[tp]
\centering
\subfloat[The large-scale industry dataset.\label{Pearson-Train-Test}]{
\begin{tabular}{|l||r|r|r|r|r||r|}
\hline
               \bf Domains & \multicolumn{1}{c|}{\textbf{PSY}} & \multicolumn{1}{c|}{\textbf{SOC}} & \multicolumn{1}{c|}{\textbf{COM}} & \multicolumn{1}{c|}{\textbf{GOV}} & \multicolumn{1}{c||}{\textbf{CRI}} & \multicolumn{1}{c|}{\textbf{Total}} \\ \hline\hline
\textbf{Train} & 12,317                                 & 15,038                               & 9,952                                & 14,151                                & 15,331                                     & 66,789                                  \\ \hline
\textbf{Test}  & 4,141                                  & 4,654                                & 3,034                                & 4,415                                 & 4,808                                      & 21,052                                  \\ \hline
\end{tabular}
}

\subfloat[SemEval-2013 dataset\label{SB-Train-Test}]{
\begin{tabular}{|l||r|r|r|r|r|r|r|r|r|r|r|r||r|}
\hline
 \bf Domains             & \multicolumn{1}{c|}{\textbf{II}} & \multicolumn{1}{c|}{\textbf{ST}} & \multicolumn{1}{c|}{\textbf{SE}} & \multicolumn{1}{c|}{\textbf{PS}} & \multicolumn{1}{c|}{\textbf{LP}} & \multicolumn{1}{c|}{\textbf{MS}} & \multicolumn{1}{c|}{\textbf{EM}} & \multicolumn{1}{c|}{\textbf{FN}} & \multicolumn{1}{c|}{\textbf{ME}} & \multicolumn{1}{c|}{\textbf{LF}} & \multicolumn{1}{c|}{\textbf{MX}} & \multicolumn{1}{c||}{\textbf{VB}} & \multicolumn{1}{c|}{\textbf{Total}} \\ \hline\hline
\textbf{Train}    & 213                              & 283                              & 539                              & 545                              & 70                               & 252                              & 430                              & 323                              & 828                              & 393                              & 697                              & 396                              & 4,969                                \\ \hline
\textbf{Test} & 24                               & 32                               & 60                               & 44                               & 8                                & 28                               & 48                               & 36                               & 92                               & 44                               & 80                               & 44                               & 540                                 \\ \hline
\end{tabular}}
\vspace{-8pt}
\caption{Domain-wise train and test splits of (a) the proprietary large-scale industry dataset and (b) SemEval-2013 dataset.\label{tab:datacombined}}
\end{table*}
\begin{table*}
\footnotesize
\begin{tabular}{|c|p{7cm}|p{5.8cm}|l|}
\hline
\bf Domain & \multicolumn{1}{c|}{\bf Question and Reference Answer} & \multicolumn{1}{c|}{\bf Student Answer} & \multicolumn{1}{c|}{\bf Label}  \\ \hline
\multirow{6}{*}{\bf PSY} & \multirow{6}{*}{\parbox[t][][t]{7cm}{Q: How does retirement affect relationships? \\ R: Retirement can cause issues as older adult couples are forced to rearrange home dynamics.}} & Retirement affects relationships in a way where there might need to be a renegotiation on what happens at home.  & correct \\\cline{3-4}
 &  & they may lose touch with the people they have formed work relationships with & partial \\\cline{3-4}
 &  & it can cause one to isolate & incorrect \\\hline
 
\multirow{3}{*}{\bf SOC} & \multirow{3}{*}{\parbox[t][][t]{7cm}{Q: What is one component of the state that makes laws? \\ R: The government legislature is one component of the state that makes laws.} } & The legislative branch (Congress) makes laws. & correct \\ \cline{3-4}
 &  & branches and state senators & partial \\\cline{3-4}
 &  & The courts and the legal systems & incorrect \\\hline

\multirow{6}{*}{\bf COM} & \multirow{6}{*}{\parbox[t][][t]{7cm}{ Q: How is attribution of a source treated with common knowledge?\\ R: Common knowledge, which is widely known information in the public domain does not need to be cited, but when in doubt whether information is common knowledge, cite it. }} & There is no need to attribute common knowledge to a source. & correct \\\cline{3-4}
 &  & that everyone must be familiar with the cited source &  partial \\\cline{3-4}
 &  & things which are of common knowledge are treated in such a way that the author is credited for his or her work & incorrect \\\hline
 
\multirow{3}{*}{\bf GOV} & \multirow{3}{*}{\parbox[t][][t]{7cm}{ Q: What does the national government share with the lower levels of government in federalism? \\ R: In federalism, the national government shares funds and information with lower levels of government.}} & regulations, transfer of funds and sharing of information & correct \\\cline{3-4}
 &  & Ability to pass laws & partial \\\cline{3-4}
 &  & Controlled by the people & incorrect \\\hline
 
 \multirow{3}{*}{\bf CRI} & \multirow{3}{*}{\parbox[t][][t]{7cm}{Q: How is crime defined?\\ R: Crime is any behavior that violates the law.}}  & The breaking of a law & correct \\\cline{3-4}
 &  & Crime is defined as deviant forms of behavior. & partial \\\cline{3-4}
 &  & deviant forms of behavior that are abnormal & incorrect \\\hline
\end{tabular} 

\caption{Some examples of questions, reference answers and student answers from each of the five domains of the large-scale industry dataset.\label{tab:sampleQ}}
\end{table*}

\begin{itemize}
\item \textbf{Generic Task-specific Learning (GTaL): }It consists of only the generic scorer $\mathcal{S}_g$ component of JMD-ASAG. The scores are converted to class-wise probabilities using a softmax layer on top of the scorer; i.e. $\mathcal{P}(x)=\sigma(\mathcal{S}_g(x))$, where $x\in \{D_d\}_{d=1,2,..,k}$.  This model learns only one scorer on the entire training set and captures the generic characteristics of domain-agnostic training. Note that, this architecture is same as BiLSTM+MaxPooling model employed by \citet{conneau2017supervised}; except that here the network is trained with short answer grading data itself. 

\item \textbf{Domain-specific Task-specific Learning (DTaL): }It consists of the domain-specific scorers, one for each domain. For the domain $D_d$, the class-wise probabilities are obtained as  $\mathcal{P}(x)=\sigma(\mathcal{S}_d(x))$, if $x\in D_d$. Since the samples from each domain affect the training of the corresponding domain-specific scorers only, it can be seen as a model that consists of $k$ domain-specific models, each trained and tested on a separate domain.
\end{itemize}



\end{itemize}
For the SemEval-2013 benchmarking dataset \cite{Dzikovska}, we also compare JMD-ASAG with various state-of-the-art non-neural and neural systems.

For fairness of comparison, we use the exact same batches and training parameters in GTaL, DTaL, and proposed JMD-ASAG.
All experimental results are reported in terms of accuracy, macro-averaged F1 and weighted-F1 metrics.
 We conclude with a discussion on the implementation details and a comparative study of the various training protocols for JMD-ASAG showing why algorithm \ref{alg:euclid} is proposed for training the model.

\begin{table*}[ht!]
\centering
\subfloat[The large-scale industry dataset. \label{Pearson-Domain}]{
\scriptsize
\begin{tabular}{|l||r|r|r||r|r|r|!{\vrule width 2pt}|r|r|r||r|r|r||r|r|r|}
\hline
\multirow{3}{*}{\bf Domains}  &   \multicolumn{6}{c|!{\vrule width 2pt}|}{\textbf{Transfer Learning \cite{conneau2017supervised}}}                 & \multicolumn{9}{c|}{\textbf{Task-Specific Learning}}\\\cline{2-8}\cline{8-16}
& \multicolumn{3}{c||}{\textbf{Generic (GTrL)}} 
& \multicolumn{3}{c|!{\vrule width 2pt}|}{\textbf{Domain-Specific (DTrL)}} 
& \multicolumn{3}{c||}{\textbf{Generic  (GTaL)}}                                                                       & \multicolumn{3}{c||}{\textbf{Domain-Specific (DTaL)}}                                                                      & \multicolumn{3}{c|}{\textbf{Joint Multi-Domain (JMD)}}                                                              \\ \cline{2-16}
                     & \multicolumn{1}{c|}{\textbf{Acc}} & \multicolumn{1}{c|}{\textbf{M-F1}} & \multicolumn{1}{c||}{\textbf{W-F1}} & \multicolumn{1}{c|}{\textbf{Acc}} & \multicolumn{1}{c|}{\textbf{M-F1}} & \multicolumn{1}{c|!{\vrule width 2pt}|}{\textbf{W-F1}} & \multicolumn{1}{c|}{\textbf{Acc}} & \multicolumn{1}{c|}{\textbf{M-F1}} & \multicolumn{1}{c||}{\textbf{W-F1}} & \multicolumn{1}{c|}{\textbf{Acc}} & \multicolumn{1}{c|}{\textbf{M-F1}} & \multicolumn{1}{c||}{\textbf{W-F1}} & \multicolumn{1}{c|}{\textbf{Acc}} & \multicolumn{1}{c|}{\textbf{M-F1}} & \multicolumn{1}{c|}{\textbf{W-F1}}  \\ \hline \hline
\textbf{PSY}    & 0.5670 & 0.5280 & 0.5558 & 0.6160 & 0.5859 & 0.6111 & 0.6638	&	0.6392	&	0.6641	&	0.6486	&	0.6171	&	0.6442	&	\textbf{0.6679}	&	\textbf{0.6421}	&	\textbf{0.6673}                    \\ \hline
\textbf{SOC}      & 0.6069 & 0.5453 & 0.5878 & 0.6432 & 0.6031 & 0.6369 & 0.6886	&	0.6461	&	0.6810	&	0.6991	&	0.6628	&	0.6944	&	\textbf{0.7073}	&	\textbf{0.6685}	&	\textbf{0.7008}                    \\ \hline
\textbf{COM}       & 0.7096 & 0.4747 & 0.6649 & 0.7452 & 0.5555 & 0.7180 & 0.7637	&	0.5642	&	0.7333	&	0.7769	&	0.6145	&	0.7571	&	\textbf{0.7844}	&	\textbf{0.6214}	&	\textbf{0.7651}                    \\ \hline
\textbf{GOV}     & 0.6539 & 0.5222 & 0.6224 & 0.6752 & 0.5717 & 0.6563 & 0.7153	&	0.6046	&	0.6928	&	0.7184	&	0.6234	&	0.7018	&	\textbf{0.7230}	&	\textbf{0.6374}	&	\textbf{0.7135}                    \\ \hline
\textbf{CRI}   & 0.6468 & 0.5527 & 0.6236 & 0.6895 & 0.6101 & 0.6751 & 0.7525	&	0.6876	&	0.7447	&	0.7606	&	0.6981	&	0.7530	&	\textbf{0.7693}	&	\textbf{0.7098}	&	\textbf{0.7631}                    \\ \hline\hline
\textbf{Overall}    & 0.6328 & 0.5440 & 0.6105 & 0.6698 & 0.6010 & 0.6583 & 0.7147	&	0.6529	&	0.7066	&	0.7185	&	0.6565	&	0.7096	&	\textbf{0.7281}	&	\textbf{0.6703}	&	\textbf{0.7216}                    \\ \hline

\end{tabular}
}

\subfloat[2-way, 3-way, and 5-way classification tasks of SemEval-2013 SciEntsBank dataset.\label{SB-UA-Domain}]{
\scriptsize
\quad\begin{tabular}{|l||r|r|r||r|r|r|!{\vrule width 2pt}|r|r|r||r|r|r||r|r|r|}
\hline
\multirow{3}{*}{}  &   \multicolumn{6}{c|!{\vrule width 2pt}|}{\textbf{Transfer Learning \cite{conneau2017supervised}}}                 & \multicolumn{9}{c|}{\textbf{Task-Specific Learning}}\\\cline{2-8}\cline{8-16}
& \multicolumn{3}{c||}{\textbf{Generic (GTrL)}} 
& \multicolumn{3}{c|!{\vrule width 2pt}|}{\textbf{Domain-Specific (DTrL)}} 
& \multicolumn{3}{c||}{\textbf{Generic  (GTaL)}}                                                                       & \multicolumn{3}{c||}{\textbf{Domain-Specific (DTaL)}}                                                                      & \multicolumn{3}{c|}{\textbf{Joint Multi-Domain (JMD)}}   
\\ \cline{2-16}
                     & \multicolumn{1}{c|}{\textbf{Acc}} & \multicolumn{1}{c|}{\textbf{M-F1}} & \multicolumn{1}{c||}{\textbf{W-F1}} & \multicolumn{1}{c|}{\textbf{Acc}} & \multicolumn{1}{c|}{\textbf{M-F1}} & \multicolumn{1}{c|!{\vrule width 2pt}|}{\textbf{W-F1}} & \multicolumn{1}{c|}{\textbf{Acc}} & \multicolumn{1}{c|}{\textbf{M-F1}} & \multicolumn{1}{c||}{\textbf{W-F1}} & \multicolumn{1}{c|}{\textbf{Acc}} & \multicolumn{1}{c|}{\textbf{M-F1}} & \multicolumn{1}{c||}{\textbf{W-F1}} & \multicolumn{1}{c|}{\textbf{Acc}} & \multicolumn{1}{c|}{\textbf{M-F1}} & \multicolumn{1}{c|}{\textbf{W-F1}}  \\ \hline \hline
\textbf{2-way} & 0.7463 & 0.7410 & 0.7461 & 0.7574 & 0.7493 & 0.7555 & 0.7815	&	0.7768	&	0.7812	&	0.7870	&	0.7805	&	0.7857	&	\textbf{0.8037}	&	\textbf{0.7986}	&	\textbf{0.8030}                   \\ \hline
\textbf{3-way} & 0.6963 & 0.6428 & 0.6916 & 0.6870 & 0.6227 & 0.6802 & 0.7352	&	0.6711	&	0.7314	&	0.7389	&	0.6899	&	0.7345	&	\textbf{0.7462}	&	\textbf{0.7111}	&	\textbf{0.7442}                   \\ \hline
\textbf{5-way} & 0.6018 & 0.5616 & 0.5996 & 0.6130 & 0.5775 & 0.6107 & 0.6387	&	0.6090	&	0.6424	&	0.6257	&	0.6057	&	0.6311	&	\textbf{0.6518}	&	\textbf{0.6252}	&	\textbf{0.6565}                   \\ \hline
\end{tabular}}
\vspace{-8pt}
\caption{Comparison of Joint Multi-Domain ASAG (JMD-ASAG) with Generic Transfer Learning (GTrL), Domain-specific Transfer Learning (DTrL), Generic Task-specific Learning (GTaL) and Domain-specific Task-specific Learning (DTaL) models on (a) the proprietary large-sclae industry dataset, and (b) 2-way, 3-way and 5-way classification tasks of SemEval-2013 SciEntsBank dataset \label{tab:resultscombined}}
\end{table*}

\subsection{Large-scale Industry Dataset}

The proprietary industry dataset contains 87K tuples of question, reference answer, student answer, and class label (grade) provided by experts.
It consists of 5 domains - Psychology (PSY), Sociology (SOC), Communications (COM), American Government (GOV), and Criminology (CRI).  
Given a question, a reference answer and a student answer, we address a 3-way classification problem involving \texttt{correct}, \texttt{partially correct}, and \texttt{incorrect} classes. 

\begin{figure}[tp]
\centering
\includegraphics[width=\linewidth]{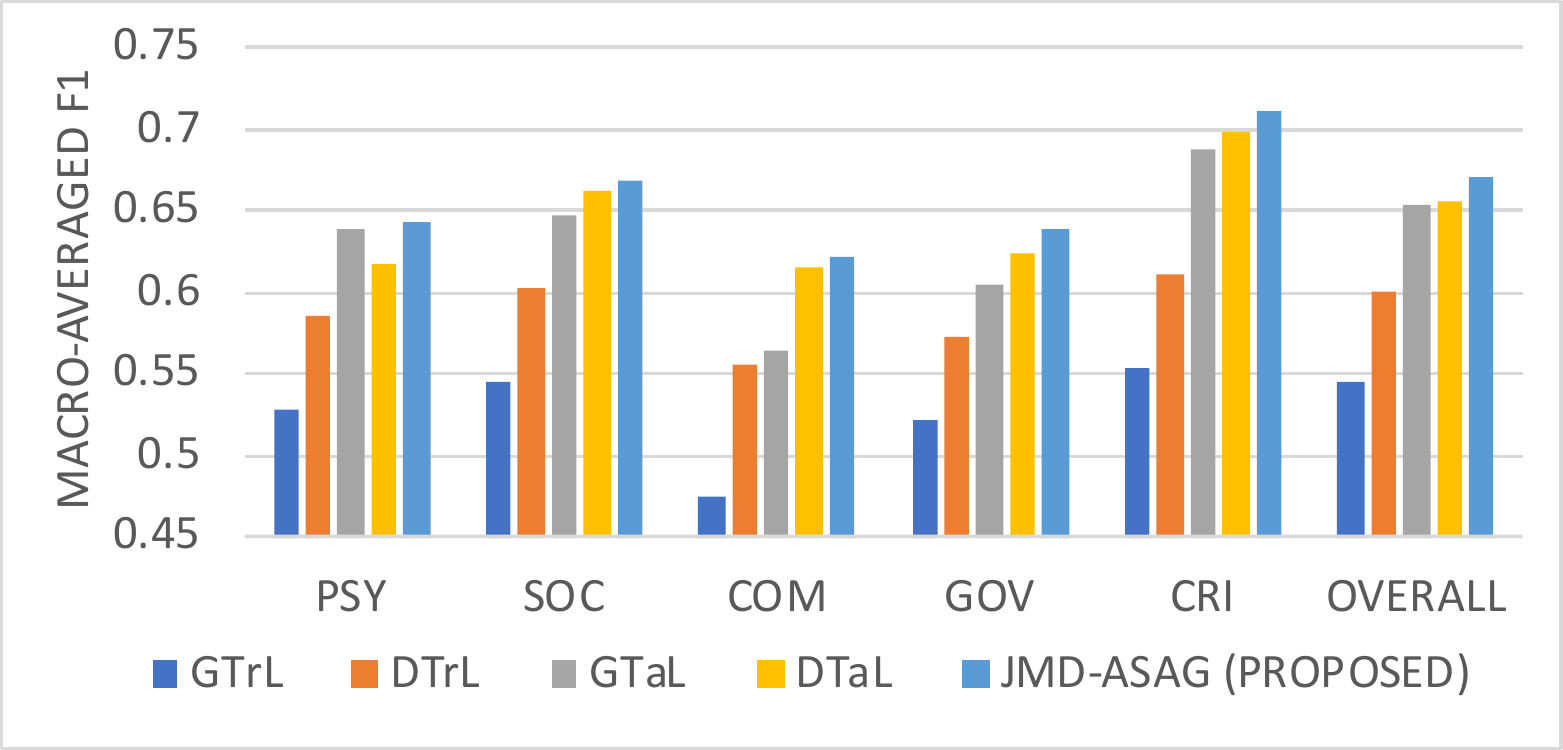}
\caption{\label{jmd-graph} Comparison of macro-averaged F1 of various models on each and combination of all domains in the industry dataset.}
\end{figure}

For each of the domains, we perform 80-20\% split of the student answers per question. They are combined for all questions to create the train and test sets. Table \ref{tab:datacombined}a shows the domain-wise train and test splits. Table \ref{tab:sampleQ} shows some examples of the questions, reference answers, student answers and class labels from all 5 domains of the large-scale industry dataset.
Based on the results reported in Table \ref{tab:resultscombined}a, following are some of our key observations.
\begin{itemize}
\item \textbf{Limitations of GTrL:} We find that GTrL exhibits significantly poor results compared to all the other models. On the overall test set, its macro-F1 is $11\%$ worse than GTaL. This is partly attributed to the Out Of Vocabulary (OOV) issue. The word embedding dictionary contains 840B words overall and out of the 46K vocabulary of the proprietary dataset, embeddings are found for only 24K terms. The task-specific models alleviate this issue by initializing all OOV words with different random embeddings and then learning them for the task. 
\item \textbf{Effect of Domains:} 
Unsurprisingly, the domain-specific characteristics are better learned and preserved when the model is trained on only one domain's data.
\begin{itemize}
\item \textbf{On Transfer Learning (GTrL vs DTrL):} All domains combined, domain-specific transfer learning yields about 6\% of macro-F1 improvement, while also consistently improving the results for each domain individually. Unsurprisingly, the domain-specific characteristics are better learned and preserved when the transferred features are trained on only one domain's data.
\item \textbf{On Task-Specific Learning (GTaL vs DTaL):} In all the domains, except for PSY, we find that DTaL shows better performance than GTaL. This is similar to the observation in transfer learning models -- domain-specific training preserves the corresponding characteristics better.
\end{itemize}

\item \textbf{Task-Specific Learning vs Transfer Learning:}
Consistently, it is observed that task-specific learning outperforms the transfer learning models within similar settings. 
\begin{itemize}
\item \textbf{Generic (GTrL vs GTaL):} When training on the combined training data, task-specific learning shows 8-13\% better macro-F1 compared to transfer learning.
\item \textbf{Domain-specific (DTrL vs DTaL):} Similarly, when there are separate models for each domain, improvements of 3-7\% are observed by virtue of task-specific learning.
\end{itemize}
These improvements suggest that task-specific learning on sufficient training data can outperform (universal) transfer learning methods.


\item  \textbf{Effectiveness of Joint Multi-Domain Learning:} JMD-ASAG illustrates the complementary benefits of GTaL and DTaL by showing significant improvements across all the domains. Compared to DTaL, the improvements in macro-F1 are mostly around 1\% in all the domains. Overall, on the combined test set of 21,052 samples, JMD-ASAG achieves about 1.5\% better macro-F1 compared to GTaL and DTaL.

Finally, we make the observation that irrespective of the specific characteristics of each domain, the performances of these models mostly follow an order - GTrL $<$ DTrL $<$ GTaL $<$ DTaL $<$ JMD-ASAG.  Figure \ref{jmd-graph} illustrates this observation.
\end{itemize}

\subsection{SemEval-2013 \cite{Dzikovska} Dataset }
This benchmarking dataset was released as part of the SemEval-2013 Shared Task 7 on ``The Joint Student Response Analysis and 8th Recognizing Textual Entailment Challenge". It consists of two different subsets - (1) Beetle, containing student responses from interaction with a dialog based tutor and (2) SciEntsBank, containing student responses to science questions. In this work, we show results only on SciEntsBank as each Beetle question contains multiple reference answers. We plan to adapt our architecture for multiple reference answers as part of the future work.
The SciEntsBank corpus consists of questions belonging to 12 science domains and their train, test splits are shown in Table \ref{tab:datacombined}b \footnote{The dataset does not provide the exact names of the domains.}. For the same set of samples, the task is performed at three different levels of granularity - (1) 2-way classification into \texttt{correct} and \texttt{incorrect} classes, (2) 3-way classification into \texttt{correct}, \texttt{incorrect} and \texttt{contradictory} classes, and (3) 5-way classification into \texttt{correct, partially correct}, \texttt{contradictory}, \texttt{irrelevant} and \texttt{non domain} classes. 
Note that the test set has the same samples across all the tasks. However, their labels change as the task becomes more granular.
Table \ref{tab:resultscombined}b shows the results pertaining to the three classification tasks\footnote{For 5-way, the macro-F1 is reported over 4 classes since the \textit{non domain} class is highly under-represented. This follows all previously published works on this dataset.}.  Following are some of the key observations.

\begin{itemize}
\item \textbf{Limitations of GTrL:} Even when the task-specific training data is significantly lesser (4,969 samples in this dataset), GTrL's macro-average F1 is up to $4\%$ worse than GTaL and DTaL. It suggests that there is a significant scope for improvement. 

\item \textbf{Effect of Domains:}
We observe moderate evidence that domain-specific training can improve learning in case of the SemEval dataset. 
DTaL is at max $1\%$ better than GTaL. Similarly, there is limited evidence of transfer learning benefiting consistently from domain-specific training.
Note that, as shown in Table \ref{SB-Train-Test}, the training samples per domain range between 70 to 697; which may be too few for effective (task-specific or transfer) learning per domain.

\item \textbf{Task-Specific Learning vs Transfer Learning:} In this dataset too, task-specific models outperform transfer learning models.
\begin{itemize}
\item \textbf{Generic (GTrL vs GTaL):} It is observed that for generic setting, task-specific learning  yields about 3-4\% higher macro-averaged F1 compared to transfer learning. Thus, training from very limited task-specific data (5K samples) can yield superior models than transfer learning from massive inference corpus (430K samples).
\item \textbf{Domain-specific (DTrL vs DTaL):} In domain-specific setting, task-specific models are around 3-6\%  better macro-F1 than those from transfer learning. As noted earlier, the domain specific data in SemEval dataset is very small, however, the task-specific learning is still more effective than transfer learning.

\end{itemize}

\item \textbf{Effectiveness of JMD-ASAG:} 
JMD-ASAG improves upon both GTaL and DTaL. For 2-way, it obtains almost 2\% better macro-averaged F1. The improvement for 3-way is even higher - 4\% and 3\% over G-ASAG and D-ASAG respectively. Finally, 5-way results are also significantly better with 2\% better macro-F1. This suggests that proposed JMD-ASAG can consistently outperform generic and domain-specific learning by incorporating benefits from both.
Table \ref{tab:example} shows two examples from this dataset where JMD-ASAG is able to predict that the student answers are correct, while GTaL and DTaL individually cannot. We believe this is owing to our model's ability to capture generic and domain-specific characteristics simultaneously.

\end{itemize}

\begin{table*}[ht!]
\centering
{
\begin{tabular}{|l||r|r|r||r|r|r||r|r|r|}
\hline
\multirow{2}{*}{Approaches}                        & \multicolumn{3}{c||}{\textbf{2-way}}                                                                          & \multicolumn{3}{c||}{\textbf{3-way}}                                                                          & \multicolumn{3}{c|}{\textbf{5-way}}                                                                          \\ \cline{2-10}
                        & \multicolumn{1}{c|}{\textbf{Acc}} & \multicolumn{1}{c|}{\textbf{M-F1}} & \multicolumn{1}{c||}{\textbf{W-F1}} & \multicolumn{1}{c|}{\textbf{Acc}} & \multicolumn{1}{c|}{\textbf{M-F1}} & \multicolumn{1}{c||}{\textbf{W-F1}} & \multicolumn{1}{c|}{\textbf{Acc}} & \multicolumn{1}{c|}{\textbf{M-F1}} & \multicolumn{1}{c|}{\textbf{W-F1}} \\ \hline \hline
\multicolumn{10}{c}{\textbf{Non-Neural Approaches}}                        \\\hline\hline
{CoMeT} \cite{ott2013comet}         & 0.7740                            & 0.7680                             & 0.7730                             & 0.7130                            & 0.6400                             & 0.7070                             & 0.6000                            & 0.5510                             & 0.5980                             \\ \hline
{ETS}  \cite{heilman2013ets}          & 0.7760                            & 0.7620                             & 0.7700                             & 0.7200                            & 0.6470                             & 0.7080                             & 0.6430                            & 0.5980                             & 0.6400                             \\ \hline
{SOFTCAR} \cite{jimenez2013softcardinality}       & 0.7240                            & 0.7150                             & 0.7220                             & 0.6590                            & 0.5550                             & 0.6470                             & 0.5440                            & 0.4740                             & 0.5370                             \\ \hline
{\citet{SultanSS16}}  & \multicolumn{1}{c|}{-}                                 & \multicolumn{1}{c|}{-}                                  & \multicolumn{1}{c||}{-}                                  & \multicolumn{1}{c|}{-}                                 & \multicolumn{1}{c|}{-}                                  & \multicolumn{1}{c||}{-}                                  & \multicolumn{1}{c|}{-}                                 & \multicolumn{1}{c|}{-}                                  & 0.5820                             \\ \hline \hline
\multicolumn{10}{c}{\textbf{Neural Approaches}}                        \\\hline\hline
{\citet{taghipour2016neural}--Best}$^\dagger$  & \multicolumn{1}{c|}{-}                                 & \multicolumn{1}{c|}{-}                                  & 0.6700                             & \multicolumn{1}{c|}{-}                                 & \multicolumn{1}{c|}{-}                                  & \multicolumn{1}{c||}{-}                                  & \multicolumn{1}{c|}{-}                                 & \multicolumn{1}{c|}{-}                                  & 0.5210                             \\ \hline
{\citet{taghipour2016neural}--Tuned}$^\dagger$ & \multicolumn{1}{c|}{-}                                 & \multicolumn{1}{c|}{-}                                  & 0.7120                             & \multicolumn{1}{c|}{-}                                 & \multicolumn{1}{c|}{-}                                  & \multicolumn{1}{c||}{-}                                  & \multicolumn{1}{c|}{-}                                 & \multicolumn{1}{c|}{-}                                  & 0.5330                             \\ \hline
InferSent \cite{conneau2017supervised} & 0.7463 & 0.7410 & 0.7461 & 0.6963 & 0.6428 & 0.6916 & 0.6018 & 0.5616 & 0.5996
\\ \hline
\citet{Saha2018}           & 0.7926                            & 0.7858                             & 0.7910                             & 0.7185                            & 0.6662                             & 0.7143                             & 0.6444                            & 0.6010                             & 0.6420                             \\ \hline \hline
\textbf{Joint Multi-Domain - ASAG}           & \textbf{0.8037}                   & \textbf{0.7986}                    & \textbf{0.8030}                    & \textbf{0.7462}                   & \textbf{0.7111}                    & \textbf{0.7442}                    & \textbf{0.6518}                   & \textbf{0.6252}                    & \textbf{0.6565}                    \\ \hline
\end{tabular}}
\caption{ Comparison of JMD-ASAG with state-of-the-art non-neural and neural models on SemEval-2013 SciEntsBank dataset. JMD-ASAG outperforms all existing models on this dataset. $^\dagger$Results as reported by \citet{riordan2017investigating}.}
\label{SB-SOA}
\end{table*}

\subsubsection{Comparison with State-of-the-Art:}
We compare JMD-ASAG with eight state-of-the-art models for ASAG. These include four non-neural models and three neural models. The non-neural models are CoMeT \cite{ott2013comet}, ETS \cite{heilman2013ets},  SoftCardinality \cite{jimenez2013softcardinality} and \citet{SultanSS16}. CoMeT, ETS and SoftCardinality are three of the best performing systems in the SemEval-2013 task.
Note that ETS \cite{heilman2013ets} is the only work of domain adaptation for ASAG and they do so by feature augmentation \cite{daume2007frustratingly}.
\citet{SultanSS16} is a more recent work on ASAG that utilizes alignment, term-weighting and vector similarity features to solve the problem.

One of the three neural models is a state-of-the-art essay scoring model by \citet{taghipour2016neural}. We use two configurations of their model for comparison - (1) best parameter set used by \citet{taghipour2016neural} and (2) tuned parameter set used by \citet{riordan2017investigating} for ASAG. The other two neural models are InferSent \cite{conneau2017supervised}, the generic transfer learning model and one model by \citet{Saha2018} that combines hand-crafted and deep learning features. Notably, \citet{Saha2018} utilizes hand-crafted token features along with deep learning embeddings, suggesting that such fusion is helpful for ASAG. Table \ref{SB-SOA} reports all the results. 


We find that JMD-ASAG yields significantly better results than all compared systems in all the three tasks. 
We report 1\% better macro-averaged F1 than \citet{Saha2018} in 2-way. The improvement in 3-way is significantly higher, with 5\% better macro-averaged F1 than \citet{Saha2018}. For 5-way, the gain is 2\%. Much to our surprise, none of the existing systems use the domain information on this dataset, which accounts for most of the improvement. We also find it particularly creditable that our end-to-end neural architecture is able to significantly outperform Saha \etal \cite{Saha2018} which combines hand-crafted features with deep learning features. As has been shown in, embedding hand-crafted features in any deep learning architecture can further enhance the performance of any short answer grading task. We leave this as part of the future work.

\subsection{Implementation Details}
We use Keras with Tensorflow as back-end for implementing our models. For the text encoder, the maximum length of the answers is set to 50 words. The embedding dimension of the words is set to 300. All word vectors are initialized with GloVe embeddings \cite{pennington2014glove} and are further updated for our task. The size of the LSTM hidden units is set to 100. The batch size is kept as 32. All models are trained for 15 epochs using categorical cross-entropy loss and Adam optimizer with a learning rate of 0.001.

\subsection{Comparison of Training Protocols}
We explore different ways of training JMD-ASAG and empirically show why algorithm \ref{alg:euclid} is the proposed way of training JMD-ASAG. We compare the following three approaches - (1) train the network such that the domain is changed after each batch, (2) train the network such that the domain is changed after each epoch, and (3) train the network such that the domain is changed only after the network has converged for the previous domain. Note that the first approach is same as algorithm \ref{alg:euclid}. The second approach is also similar but with lines 5 (the loop of \texttt{batches}) and 6 (the loop of \texttt{domains}) in algorithm \ref{alg:euclid} interchanged. In the third approach, the loop that iterates over \texttt{domains} (line 6 in algorithm \ref{alg:euclid}) comes before the other two loops. Table \ref{Pearson-Training} compares the three approaches on the combined test set of the industry dataset.

\begin{figure}
\includegraphics[width=\linewidth]{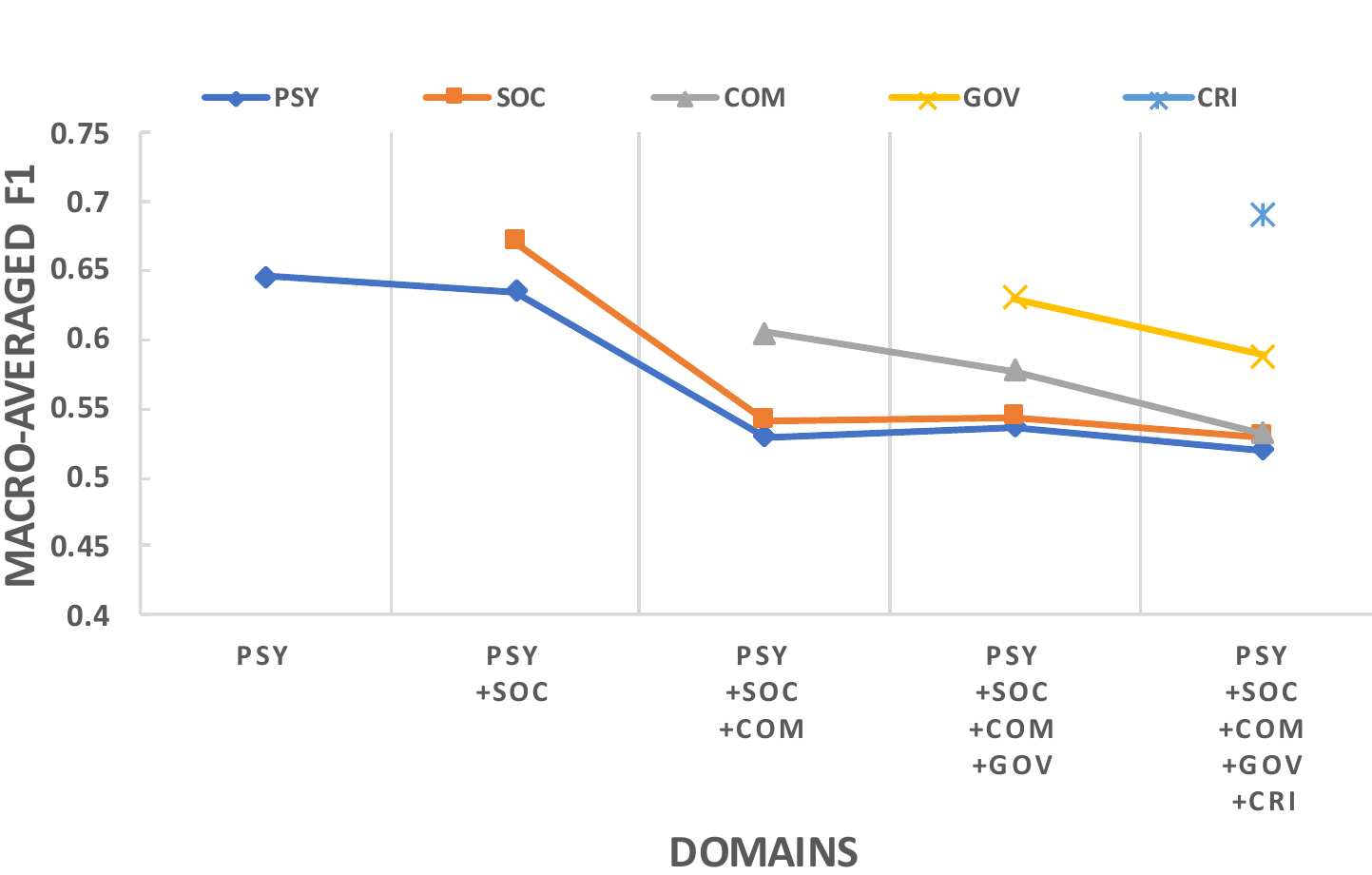}
\caption{ Training on new domains results in successive decrease in performance of previously seen domains.}\label{fig:2}
\end{figure}

Batch- and epoch-wise trained models show similar performances and massively outperform domain-wise trained models. This however, is unsurprising. Whenever the model is trained on a particular domain's data until convergence, it is fine-tuned for the current domain, and subsequently decreases the performance on the previous domains. 
This leads to a progressive reduction in numbers for each of the previous domains and eventually, lowering the performance on the overall test set. 
This phenomenon is observed in Figure \ref{fig:2}. On training with a new domain (horizontal-axis), the macro-F1 (vertical-axis) for all the previous domains keep decreasing progressively.
\begin{table}
\centering
\begin{tabular}{|l||c|c|c|}
\hline
                 & \textbf{Acc}    & \textbf{M-F1}   & \textbf{W-F1}   \\ \hline\hline
\textbf{Batch}   & 0.7281          & \textbf{0.6703} & \textbf{0.7216} \\ \hline
\textbf{Epoch}   & \textbf{0.7297} & 0.6700          & 0.7211          \\ \hline
\textbf{Domain} & 0.6784          & 0.5871          & 0.6526          \\ \hline
\end{tabular}
\caption{ Comparison of various training protocols of JMD-ASAG on the industry dataset.}
\label{Pearson-Training}
\end{table}
 
\section{Conclusion and Future Works}
Till date, one of the fundamental challenges towards building a real-world deployable intelligent tutoring system has been the lack of adaptability of an automatic short answer grading across various domains or subjects. While almost all existing works have modeled the problem as a typical textual similarity problem independent of the domain, we find that in ASAG the notion of similarity varies across domains. In response, we propose JMD-ASAG, a novel neural network architecture for joint multi-domain learning of ASAG. JMD-ASAG not only learns the domain-specific characteristics of similarity but also the generic aspects that is universal to the properties of the language. For $k$ domains, JMD-ASAG achieves both these by learning $k$ domain-specific similarity scorers and one generic scorer in an end-to-end trainable neural architecture. Also, it does not rely on a large corpus for learning the generic characteristics. Empirical evaluation on a proprietary large-scale industry dataset and a benchmarking dataset show that JMD-ASAG outperforms a state-of-the-art transfer learning model and models that only employ generic or domain-specific learning from task-specific training data. We report state-of-the-art results on the benchmarking dataset and also empirically show why our proposed algorithm for training the model is the most optimal among various other protocols. We believe JMD-ASAG can further benefit from better similarity scorers; exploring this is left as part of the future work.

In the quest for building a first of its kind large-scale intelligent tutoring system, we have deployed our JMD-ASAG model trained on the five domains of the industry dataset. 
The pilot study of the system is currently being carried out with about thousand students across the globe.
In the future, we plan to scale our system to 100 subjects. Our architecture is simple yet effective, ensuring that such scale up should be trivial. Also, we believe that with increased number of domains, the generic characteristics of the language will be better learned, leading to further gains in performance. Finally, although our results are specific to the task of ASAG, we believe that the architecture of JMD-ASAG can be directly applied to any semantic similarity task that requires capturing generic and domain-specific characteristics. We plan to explore this too as part of the future work.


\bibliographystyle{ACM-Reference-Format}
\bibliography{sample-bibliography}

\end{document}